\documentclass[10pt, a4paper, twocolumn, teaser, showabstract]{naverlabseurope}
\usepackage{booktabs}
\usepackage{CJKutf8}
\usepackage{lipsum}
\usepackage{multicol}
\usepackage{tikz,tkz-kiviat,pgfplots}
\usepackage{bbding}
\usepackage{pifont}
\usepackage{multirow}
\usepackage{latexsym}
\usepackage{tabularx}
\usepackage{seqsplit}

\newcommand{\ours}{NAVER LABS Europe Submission to the Instruction-following 2026 Short Track}

\newcommand{\fix}[1]{\textcolor{red}{ #1}}
\newcommand{\seamless}{\texttt{SeamlessM4T-v2-large}}
\newcommand{\llama}{\texttt{Llama-3.1-8B-Instruct}}
\newcommand{\qwen}{\texttt{Qwen3-4B-Instruct-2507}}
\newcommand{\speechmapper}{\texttt{SpeechMapper}}

\graphicspath{{figures/}}

\title{NAVER LABS Europe Submission to the Instruction-following 2026 Short Track}
\titlerunning{\ours}

\authors{Marcely Zanon Boito$^{1}$, Hemant Yadav$^{2}$, Jean-Luc Meunier$^1$, Ioan Calapodescu$^{1}$}
\affiliations{$^1$NAVER LABS Europe, France\\
$^2$IIIT Delhi, India}
\correspondingauthor{marcely.zanon-boito@naverlabs.com}

\teaserfig{\includegraphics[width=0.95\textwidth]{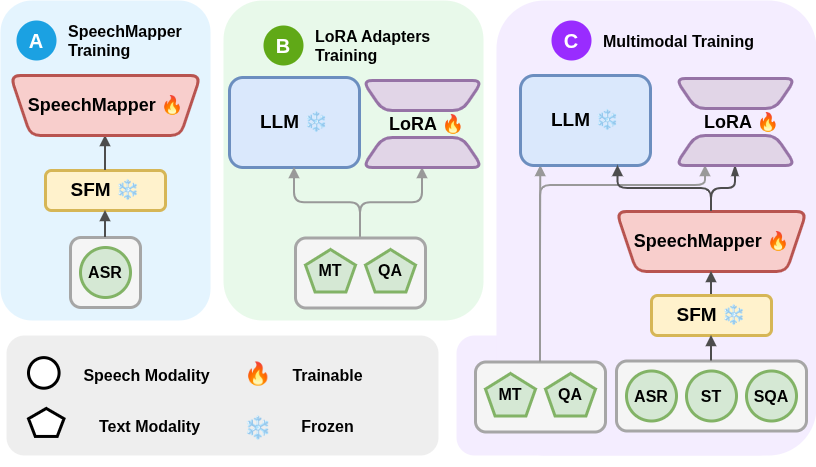}}

\teasercaption{Our training pipeline. A speech projector (SpeechMapper)~(A) and text LoRA adapters~(B) are trained in parallel using ASR and text-to-text data, respectively. These modules are then integrated during a brief multimodal adaptation step~(C).\label{fig_apx:main_figure}}

\begin{abstract}
In this paper, we describe NAVER LABS Europe's submission to the instruction-following speech processing short track at IWSLT 2026. We participate again in the constrained setting, developing systems capable of jointly performing ASR, ST, and SQA from English speech into Chinese, Italian, and German.
Building on our previous submission, ranked first in last year's short track, we update our multi-stage training pipeline by replacing the speech projector with \speechmapper, a method for learning a speech-to-LLM embedding projector using only ASR data. In addition, we introduce a synthetic SQA dataset, \textit{fakACL}, composed of artificially generated scientific presentations. This dataset is built by prompting the LLM backbone, segmenting the generated talks, and synthesizing speech with \seamless. The combination of an improved speech projection mechanism and domain-specific synthetic data allows our model to outperform last year's best short-track system, while being considerably more compact and relying on a weaker LLM backbone. This year's results place our system tied for first place in the overall short track ranking.
\end{abstract}

\begin{document}
\contributions{}
\maketitle

\section{Introduction}

Fueled by the progress of text-only LLMs, a growing number of speech-based assistants have recently been proposed, addressing both semantic~\cite{rubenstein2023audiopalm,tang2023salmonn,defossez2024moshi,hu2024wavllm,ambilduke2025tower,speechmapper} and acoustic speech tasks~\cite{huang2024audiogpt,nguyen2024spirit,xu2025qwen3}. In this context, the second edition of the \textit{IWSLT Instruction-following Speech Processing Track}~\cite{adelani-etal-2026-iwslt} provides a common benchmark for developing speech assistants capable of performing semantic tasks directly from speech.
The challenge proposes to leverage LLMs and speech foundation models~(SFMs) to build systems that can perform multilingual tasks from English speech input, guided by textual multilingual instructions. 

NAVER LABS Europe (NLE) again participates in the constrained setting of the \textit{short track}, which includes automatic speech recognition (ASR), speech translation (ST), multilingual spoken question answering (SQA), and a surprise instruction-following task revealed only at test time. The target languages for ST and SQA are Chinese, Italian, and German. The target languages for the surprise task were German and Chinese.
Participants are allowed to use the \seamless~\cite{barrault2023seamlessm4t} speech backbone and the \qwen~\cite{qwen3technicalreport} LLM for both training and data generation. Building on our previous winning submission~\cite{lee-etal-2025-naver}, we adopt a similar multi-stage training pipeline to develop a speech LLM capable of following multilingual instructions.

Our approach trains two components in parallel: (1)~a speech-to-text projector that maps averaged speech representations from the \seamless{} encoder into the embedding space of a frozen LLM, and (2)~text-only LoRA adapters~\cite{hu2022lora} applied to the same frozen LLM. These components are subsequently combined through a (3)~short SFT stage on multimodal multilingual data.
Compared to last year, we introduce two main improvements. First, we replace the speech projector with an updated version of \speechmapper~\cite{speechmapper}, which enables learning a speech-to-embedding mapping without requiring LLM forward passes, substantially lowering computational and hardware requirements. Second, we construct a synthetic scientific dataset, \textit{fakACL}, to reduce the domain gap between training and evaluation data.

This system paper is organized as follows. 
Section~\ref{sec:data} describes the preprocessing applied to the data used in this challenge. Section~\ref{sec:speechmapper} presents the updated \speechmapper{} model we leverage in this work.
Sections~\ref{sec:models} and~\ref{sec:settings} describe our training pipeline and experimental settings, respectively.
Section~\ref{sec:experiments} presents our experiments and discussion.
Section~\ref{sec:submission} presents the submitted system. Section~\ref{sec:conclusion} concludes the paper.
\section{Data}\label{sec:data}

We leverage all available data from the constrained setting for training, including CoVoST2~\cite{wang2020covost}, EuroParlST~\cite{jairsan2020a}, GigaST~\cite{ye23b_interspeech}, and LibriSQA~\cite{librisqa}. This data is also synthetically augmented to cover additional target languages via \seamless{} MT.
In addition, we introduce \textit{fakACL}, a synthetic dataset of scientific presentations generated by prompting the LLM backbone and synthesizing the outputs with \seamless{} TTS. The validation sets of EuroParlST, CoVoST2, and LibriSQA are used for model validation, while MCIF~\cite{papi2026mcif} is used for final model selection.

We now present our data preprocessing~(Section~\ref{sec:datapreproc}) and prompt format~(Section~\ref{sec:promptformat}).

\subsection{Data Preprocessing}\label{sec:datapreproc}

We create both speech-to-text and text-to-text instructions to train our systems. Appendix Table~\ref{tab:trainingdata} lists our splits and number of samples. In all cases where synthetic translation is generated from English text, this procedure is done via \seamless{} MT followed by filtering of the result using reference-free COMET\footnote{\texttt{Unbabel/wmt22-cometkiwi-da}}~\cite{rei-etal-2022-comet} to filter out all translations that did not score at least 0.85. Below we detail dataset-specific preprocessing.

\paragraph{GigaST, CoVoST2 and EuroParlST} CoVoST2 and GigaST cover English to German and Simplified Chinese language directions. EuroParlST covers English to German and Italian. ASR splits for CoVoST2 and EuroParlST datasets were built by merging the existing language splits and de-duplicating the audio files. For all, language-specific ST and MT splits were created by aligning translations to English speech and reference transcriptions, respectively. For GigaST, and because our LLM backbone is particularly weak in Italian~(see Table~\ref{tab:results1}), we produce additional synthetic Italian translations. 

\paragraph{LibriSQA}
LibriSQA is built on the 300-hour split of LibriSpeech~\cite{panayotov2015librispeech}. The dataset is divided in \textit{PartI} and \textit{PartII}, which share the same speech content, but differ in the format of the questions and answers. \textit{PartI} consists of direct questions to be the answered by the system, while \textit{PartII} presents multiple-choice questions~(MCQ) corresponding to those in \textit{PartI}. We performed the following modifications to this dataset:
 \begin{itemize}
    \item \textbf{MCQ answer expansion:} For \textit{PartII}, answers are provided as the letter corresponding to the correct option. This results in a very limited number of tokens contributing to the loss, which we found to be suboptimal in past experiments. To address this, we expand the target answers to include the full option text (e.g., instead of predicting “A”, the model is trained to produce ``A \textless{}option A text\textgreater{}'').
    \item \textbf{Multilingual SQA/QA:} We translate the English text into the three target languages. For \textit{PartI}, we retain only question-answer pairs whose \textit{both} translations exceed the quality threshold. For \textit{PartII}, we translate only the questions and their associated answer options, since the correct answer is implicitly defined by the options. We then apply a parsing step to discard examples where the A–B–C–D structure is altered or lost during translation. This parser also recovers the correct answer by aligning the translated options with the original references.
    \item \textbf{MT split:} We create an MT split for each target language by exploiting the alignment between questions and answers obtained during the creation of our SQA split. We do not create a corresponding ST split, as we consider this dataset to be significantly out-of-domain for the challenge, particularly in terms of acoustic conditions, and including more of it could be detrimental during training.
    \item \textbf{Similarity-based Invalid splits:} Some questions in LibriSQA are relatively broad (e.g., ``How did the person in the text act?''), making it difficult to reliably mismatch them with unrelated audio segments. To address this, we construct mismatched pairs by selecting question–transcript combinations with low semantic similarity. We first shuffle and randomly pair transcripts and questions, then compute sentence-level representations using a Sentence Transformer\footnote{\texttt{\seqsplit{sentence-transformers/multi-qa-MiniLM-L6-cos-v1}}} to measure cosine similarity between them. We retain only pairs with low similarity scores.\footnote{Based on manual assessment, we set this to $\leq 0.1$.} This process yields 48,346 and 58,682 mismatched pairs for \textit{PartI} and \textit{PartII}, respectively. 
    \item \textbf{Qwen-based Invalid splits:} To address potential noise introduced by our similarity-based invalid splits we created, we include a more general-purpose set of invalid questions by prompting \qwen{}~(see Appendix~\ref{sec:qweninvalid} for details). After post-processing, this yields 62,040 English questions, which are then translated and filtered for quality using the same procedure described above. Finally, these questions are randomly paired with speech audio segments to construct the invalid splits.
 \end{itemize}

\paragraph{fakACL} In both this and the previous edition of the challenge, the validation data is drawn from scientific paper presentations at ACL conferences. Motivated by this, we constructed a synthetic dataset targeting this domain to reduce the mismatch between training and test conditions, and to provide additional relevant QA data, as the constrained SQA dataset is relatively distant from the test setting.
We generated fakACL ASR/ST/SQA data by first producing short presentation scripts for NLP papers using \qwen{}. These scripts were segmented into chunks of two to three sentences and synthesized into speech using \seamless{} TTS. Each segment transcript was then fed back into the LLM to generate two questions answerable from the corresponding content. Appendix~\ref{sec:fakaclcreation} provides full details of the dataset creation process.

\subsection{Prompt Format}\label{sec:promptformat}
The goal of the short track of this challenge is to produce a model that is able to 1) transcribe English speech; 2) translate English speech into Italian, German and Chinese; 3) Answer multilingual questions using English speech as input. In this setting, the language of the question must match the language of the answer. 

Last year we designed an unified prompt with consistent structure: regardless of the task (ASR, ST, or SQA), the user turn begins by encapsulating the speech embeddings within textual tags. This is followed on a new line by a task-specific instruction formulated as a question in the target language, and finally, another line containing a common suffix. Our last year results~\cite{lee-etal-2025-naver} highlighted that by forcing the instruction~(i.e. question) to be in the target language, we were able to reduce confusion between the tasks of ASR/ST, while also helping the system to generalize to languages it was not originally trained for.\footnote{For instance, in the SpeechMapper paper~\cite{speechmapper}, our IWSLT25 system generalizes to Spanish and French ST via multilingual prompting, despite the fact that its LoRA weights were never trained on these languages.} Therefore, this year we only experiment with this setting, in which the instruction is given in the target language.

We additionally include a template for multiple choice questions~(MCQ), for the surprise task, and for zero-shot generation~(LLM backbone and \speechmapper{} in projector-only mode). The list of updated templates used is available in Appendix Table~\ref{tab:prompts}.

\section{Speech Projector: \speechmapper}\label{sec:speechmapper}

In this work we use an improved version of the original \speechmapper{} projector~\cite{speechmapper}.
SpeechMapper is a speech-to-LLM embedding projector for semantic information, and training it only requires an LLM's tokenizer and its embedding layer. 
The benefits are twofold: (1)~the size of LLM does not increase GPU memory required for training;\footnote{The only increase in computational cost comes from the embedding dimensionality, which is typically not proportional to the LLM size, keeping the approach computationally efficient.} (2)~it mitigates prompt overfitting. We refer the reader to the original paper for further details.

The decision to replace last year’s simple transformer based projector with the more parameter-heavy \speechmapper{} is primarily motivated by practical considerations. In preliminary experiments, we attempted to reproduce the pipeline from~\citet{lee-etal-2025-naver}, but observed very poor performance for projector-only models (exceeding 200\% WER). We attribute this low-performance compared to last year to the lower dimensionality of the embedding space in this year's LLM. We hypothesize that projecting into a smaller embedding space requires higher precision, and therefore greater model capacity, which \speechmapper{} is better able to provide.

Below, we describe our main modification to \speechmapper{} relative to~\citet{speechmapper}: the training objective.

\subsection{New \speechmapper{} Training Objective}
Our updated version of \speechmapper{} is trained using four loss functions designed to align speech to LLM input embeddings. Speech is encoded into a sequence of embeddings $Z_s \in \mathbb{R}^{T \times d}$ using the frozen SFM and fed to the \speechmapper{}. The corresponding target sentence is processed to a sequence of LLM token embeddings $Z_t \in \mathbb{R}^{T' \times d}$ using the frozen LLM embedding layer. The resulting speech sequence length $T$ is significantly larger than the text sequence length $T'$. To mitigate this length mismatch, we pad the target embedding sequence with the LLMs \texttt{[PAD]} token embedding to match the speech sequence length~(Eq.~\ref{eq:pads}).

\begin{equation}\label{eq:pads}
Z_t = [z_t^{(1)}, \dots, z_t^{(T')}, z_{\text{pad}}, \dots, z_{\text{pad}}] \in \mathbb{R}^{T \times d}
\end{equation}

Therefore, the output of \speechmapper{} is semantic embeddings followed by pad embeddings, implicitly capturing sequence length. We optimize the following objectives.

\paragraph{L1 Alignment Loss}
The element wise L1 distance between speech and text embeddings explicitly enforce feature-level alignment between the two modalities~(Eq.~\ref{eq:l1}). This loss replaces the MSE loss from \citet{speechmapper}.

\begin{equation}\label{eq:l1}
\mathcal{L}_{\text{L1}} = \frac{1}{T d} \sum_{t=1}^{T} \left\lVert z_s^{(t)} - z_t^{(t)} \right\rVert
\end{equation}

\paragraph{Cosine Similarity Loss}
To encourage angular alignment in the embedding space, we keep the original cosine similarity loss from the original paper, and minimize the cosine distance between corresponding speech and text representations~(Eq.~\ref{eq:cos}).

\begin{equation}\label{eq:cos}
\mathcal{L}_{\text{cos}} = \frac{1}{T} \sum_{t=1}^{T} \left( 1 - 
\frac{z_s^{(t)} \cdot z_t^{(t)}}{\lVert z_s^{(t)} \rVert_2 \lVert z_t^{(t)} \rVert_2} \right)
\end{equation}

\paragraph{Softmax Contrastive Loss}
To enforce separation from non target tokens, we use softmax function to maximize similarity to the positive class embeddings and minimize similarity to all other embeddings.
Let $E \in \mathbb{R}^{V \times d}$ denotes the LLM's embedding layer, where $V$ is the vocabulary size. For each speech representation $z_s^{(t)}$, we compute cosine similarity scores with all embeddings and use these as logits, with $s_t \in \mathbb{R}^{V}$ and $y_t \in \mathbb{R}^{V}$ denoting respectively the logits and the one-hot vector corresponding to the target token at position $t$~(Eq.~\ref{eq:softmax}). The negative log likelihood encourages \speechmapper{} output embeddings to be close to its correct token embedding while remaining well separated from all other tokens in the vocabulary.

\begin{equation}\label{eq:softmax}
\mathcal{L}_{\text{softmax}} = - \frac{1}{T} \sum_{t=1}^{T} y_t^\top \log(\text{softmax}(s_t))
\end{equation}

\paragraph{Connectionist Temporal Classification (CTC)}
We found that adding CTC \cite{graves2006ctc} loss ($\mathcal{L}_{\text{ctc}}$) at an intermediate layer helps in stabilizing \speechmapper{}'s training and output embeddings quality.

\paragraph{Final Objective}
The overall training objective is a weighted sum of the losses:
\begin{equation}
\mathcal{L} = \mathcal{L}_{\text{L1}} + \mathcal{L}_{\text{cos}} + 0.1*\mathcal{L}_{\text{softmax}} + 
\mathcal{L}_{\text{ctc}}
\end{equation}

\section{Training Pipeline}\label{sec:models}

Our training pipeline is illustrated in Figure~\ref{fig_apx:main_figure}. We first train two components in parallel: (A)~a SpeechMapper speech-to-embedding projector using ASR data, and (B)~text LoRA weights using MT and QA data. These components are subsequently reloaded and jointly adapted on a mixture of speech and text tasks~(C). In this section, we describe the key components of this training pipeline.

\paragraph{Foundation Models} For speech, we leverage \seamless{}  model, extracting speech representations for all our audio data from its 24th speech encoder layer~(i.e. the last layer). Prior to training, we average every two consecutive frame vectors, reducing significantly the sequence length. We highlight a minor difference from last year’s setup: we average two consecutive frame vectors instead of three. Prior experiments with \speechmapper{} motivated this choice, indicating that a lighter compression before the projector’s CNN layers yields better performance. 
All our models are built on top of a frozen \qwen{}.

\paragraph{SpeechMapper Settings} Our \speechmapper{} follows the architecture described in~\citet{speechmapper}, comprising two consecutive blocks, each consisting of a CNN, \textit{N} self-attention layers, and a feed-forward projection to a higher dimensional space. To accommodate a CTC head at the end of the first block, we modify the convolutional strides from the original setting of 2 in both blocks to 1 and 4 for the first and second blocks, respectively. 
We train the model with $N=6$, using initial, intermediate, and output dimensionalities of $1024$, $2048$, and $2560$, respectively.

\paragraph{LoRA Adapters} LoRA adaptation~\cite{hu2022lora} is applied to both the self-attention (Q/K values, output projection) and feed-forward modules, and across all LLM layers. We use $rank=8, \alpha=16$. We do not use dropout.

\paragraph{Data Sampling Strategy} For training the models in Figure~\ref{fig_apx:main_figure} (B) and (C), we define an epoch as $X$ steps across the dataset, where $X=\frac{|\text{speech}\_\text{examples}|}{\text{batch}\_\text{size}}$. 
To construct the data for each epoch, batches are sampled by first applying the predefined task-level sampling ratios, followed by sampling according to domain-level splits within each task. 
In the multimodal training setting (speech and text tasks mixed), we consider speech as our \textit{main} modality, using it for defining epoch size and task sampling ratio. Whenever a sampled speech task and language pair has a textual counterpart (e.g., ST corresponds to MT; SQA to QA), we also sample a batch from the corresponding textual task. In practice, this results in alternating batches, such that a sampled ST en-de batch is followed by an MT en-de batch. We find that incorporating textual data in this manner consistently improves the final model performance.

\section{Experimental Settings}\label{sec:settings}

\paragraph{Codebase} We train our models using an internal fork of \texttt{torchtune}~\cite{torchtune}, which allows us to process interleaved representations of text and high-dimensional vectors within the user turn during instruction tuning. We also implement our updated version of \speechmapper{} on this codebase. For multimodal training, 
the high-dimensional vectors pass through \speechmapper{}, while the text prefix and suffix user prompts are processed by the LLM embedding layer. The obtained speech and text embeddings are both concatenated and fed into the first layer of the LLM which is trained on the masked input with standard cross-entropy loss. 
Different learning rate schedulers and optimizers are employed for \speechmapper{} and the LoRA weights, allowing for more controlled and effective training of these distinct model components.

\paragraph{Inference Settings} 
We perform inference using \texttt{torchtune}, with a batch size of 1, greedy decoding, and with the maximum number of new tokens limited to 100. This decoding strategy was consistently applied across all experimental settings.

\paragraph{Evaluation Metrics}
We evaluate our models on speech~(ASR, ST, SQA) and text~(MT, QA) tasks when relevant. For ASR, we score word error rate (WER) using HuggingFace evaluate library with default settings and MMS normalization~\cite{pratap2024scaling}. For ST/MT we score COMET \cite{rei-etal-2022-comet}.\footnote{\texttt{Unbabel/wmt22-comet-da}} For SQA/QA, we use LLM-as-a-judge evaluation scripts  from the \texttt{bergen} library\footnote{\url{https://github.com/naver/bergen}}~\cite{rau-etal-2024-bergen}. We use their ``yes/no'' quality assessment evaluation format including the reference question and answer, and generated output. We report average accuracy across three LLMs: \texttt{EuroLLM-9B-Instruct}~\cite{eurollm}, \texttt{Gemma3-12B-Instruct}, and \texttt{Gemma3-27B-Instruct}~\cite{gemma3}.

\paragraph{Baselines} We compare our results with both backbones we use for training. We evaluate MT using the reference transcripts and \qwen{} in zero-shot settings, and we evaluate \seamless{} for ASR, ST and MT. Additionally, we present results for last year's best short track system, referring to it as \textbf{BEST-IWSLT25-IF}~\cite{lee-etal-2025-naver}.

\section{Experiments}\label{sec:experiments}

\begin{table*}
\centering
\resizebox{\linewidth}{!}{
\begin{tabular}{lc|ccc|cc}
\toprule
\textbf{Model} & \textbf{ASR (WER)} & \multicolumn{3}{c}{\textbf{ST/MT (COMET)}} & \multicolumn{2}{|c}{\textbf{SQA/QA (LLM-as-judge)}} \\
 &  & en-de & en-it & en-zh & Part I & Part II \\
\midrule
\multicolumn{7}{c}{\textit{Text-only Models (zero-shot LLMs, SFM and LoRA)}} \\\midrule
\seamless{} (MT) & - & 81.4 & 86.7 & 80.8 & - & - \\
\llama{} & - & \textbf{81.9} & 84.1 & 77.0 & 86.6 & 70.8 \\
\qwen{} & - & 71.0 & 67.7 & 74.3 & 89.1 & 70.2 \\
\qwen{} + LoRA (B) & - & 80.7 & \textbf{86.9} & \textbf{84.7} & \textbf{89.9} & \textbf{82.6} \\
\midrule
\multicolumn{7}{c}{\textit{SFM and Projector-only Model}} \\\midrule
\seamless{} (ASR/ST) & \textbf{5.9} & \textbf{78.3} & 76.9 & 78.0 & - & - \\
\speechmapper{} (A) & 14.2 & 73.5 & \textbf{80.1} & \textbf{79.7} & 84.4 &	72.1 \\
\midrule
\multicolumn{7}{c}{\textit{Multimodally Trained Models}} \\\midrule
BEST-IWSLT25-IF & \textbf{7.3} & \textbf{77.3} & 84.2 & 80.2 & 82.0 & 63.0\\
\speechmapper{} + LoRA (C) setup 1 & 8.2 & 75.1 & 83.2 & 80.6 & 86.2 & \textbf{82.5}\\
\speechmapper{} + LoRA (C) setup 2 & 7.4 & 76.3 & \textbf{84.4} & \textbf{81.3} & \textbf{87.9} & 80.2 \\
\bottomrule
\end{tabular}}
\caption{Results for ASR (LibriSpeech clean/other, EuroParlST, CoVoST2), ST/MT (CoVoST2, EuroParlST), and SQA/QA (LibriSQA PartI and PartII). ASR and ST/MT scores are reported as weighted averages across datasets, proportional to the number of samples, while SQA/QA results are averaged across judges and reported as accuracy.}
\label{tab:results1}
\end{table*}
\begin{table}
\resizebox{\columnwidth}{!}{
\begin{tabular}{lccc}
\toprule
\textbf{Model}                    & \textbf{ASR}  & \textbf{ST}    & \textbf{SQA}   \\\midrule
BEST-IWSLT25-IF          & 12.6          & 0.743          & 0.417          \\
\speechmapper{} (A) & 32.2          & 0.711          & 0.225          \\
\speechmapper{} + LoRA (C) setup 1 & 12.0          & 0.772          & \textbf{0.428} \\
\speechmapper{} + LoRA (C) setup 2 & \textbf{10.5} & \textbf{0.781} & 0.400\\\bottomrule
\end{tabular}
}
\caption{Averaged MCIF scores for all speech models.}
\label{tab:mcif_results}
\end{table}

We now present our results for ASR, ST, and SQA. Section~\ref{sec:experiments:models} introduces the models used in our experiments, followed by results and discussion in Section~\ref{sec:experiments:results}.

\subsection{Our Models}\label{sec:experiments:models}

\paragraph{\speechmapper{}} We train \speechmapper{} on ASR data from CoVoST2, EuroParlST, GigaST, and LibriSQA, without applying any up-sampling. Training is performed for 500k steps using AdamW with a learning rate of $1e-4$ and 50k warm-up steps. We employ dynamic batching with gradient accumulation set to 2. As \qwen{} does not provide a dedicated padding token, we use the reserved (untrained) token $151664$ for padding. The model is trained on $4\times$A100-80GB GPUs for approximately two days. 

\paragraph{B. Text-only LoRA (MT/QA)} The LoRA weights~(B) are trained on all available real and synthetic data from Appendix Table~\ref{tab:trainingdata}, with task-level sampling ratios of 0.6 and 0.4 for MT and SQA, respectively. For MT, language sampling ratios are set to 0.4/0.3/0.3 for de/it/zh, while for QA they are 0.2/0.3/0.3/0.2 for en/de/it/zh. We train for 30k steps using AdamW with learning rate of $3e-4$, weight decay of $0.1$, and $100$ warm-up steps. Batch size of $16$, and gradient accumulation of $8$ is used. This model trains for approximately 4 days in a single A100-80GB. 

\paragraph{C. Multimodal (A + B)} We restart training by reloading both modules described above. We explore various combinations of the available datasets and obtain the best performance with the configuration detailed in Appendix Table~\ref{tab:listdatasets}. We use task-level sampling ratios of 0.3/0.4/0.3 for ASR, ST, and SQA, respectively. For ST/MT, language sampling ratios are set to 0.4/0.4/0.2 for de/it/zh. For SQA/QA, we use uniform language sampling over the four languages, assigning a ratio of 0.2 each for the valid set and 0.05 each for the invalid set. We use learning rate of $5e-5$ for the \speechmapper{}, and of $1e-5$ for the LoRA weights~(setup 1) or $5e-5$ for the LoRA weights~(setup 2). We use a batch size of 8, and gradient accumulation of 6. This model trains for 3K steps, approximately 2~hours in a single A100-80GB. 

\subsection{Results and Discussion}\label{sec:experiments:results}

Table~\ref{tab:results1} present results for ASR, ST/MT and SQA/QA for all of our models and baselines over the available test splits. For ASR, we evaluate using EuroParl, CoVoST and Librispeech clean/other. For ST/MT, we evaluate on EuroParl and CoVoST, excluding GigaST which we find to be particularly noisy. SQA/QA results cover only English, and are an average over scores obtained for the different LLM-as-judge models. Table~\ref{tab:mcif_results} present MCIF scores~(ASR, ST and multilingual SQA) obtained for the speech models. Below we discuss our main findings.

\paragraph{MT Results} Looking at the MT results for text-only models~(top portion of Table~\ref{tab:results1}), we observe that this year's backbone performs substantially worse than \seamless{} and \llama{}. This is expected, as the backbone has roughly half the parameters of last year's LLM. Results are particularly poor for Italian, motivating us to up-sample this language during fine-tuning. Finally, we also present results for our best LoRA setting, which considerably improves performance across both MT and QA tasks.

\paragraph{Projector-only Results} 
We present results for our best \speechmapper{} configuration trained to produce \qwen{} embeddings. Overall, this setting proves challenging: the LLM appears particularly sensitive to noise in the input embeddings, especially compared to larger backbones tested in \citet{speechmapper}. Our best \speechmapper{} setup performs correctly in in-domain settings~(Table~\ref{tab:results1}), but we find it to under-perform considerably in the MCIF dataset~(Table~\ref{tab:mcif_results}), where it handles poorly named entities. 
In addition, zero-shot usage of \qwen{} is difficult, as the model reorganizes and/or rephrases the transcription text in many instances, or responds in an overly verbose manner, both cases negatively affect evaluation scores. Qualitative examples illustrating these behaviors are provided in Appendix Table~\ref{tab:generationsm}.

\paragraph{Multimodal Models} We experiment with a wide range of dataset combinations and sampling strategies, and report results only for our best-performing multimodal models (bottom portion of Table~\ref{tab:results1} and Table~\ref{tab:mcif_results}). These models are able to match the performance of last year’s system, despite relying on a smaller LLM backbone.
Overall, we find that improving performance jointly on ASR and SQA is more challenging this year, as gains in one task often come at the expense of the other. This trade-off is illustrated by Setups 1 and 2: while Setup 1 achieves better results on LibriSQA Part II (MCQ) and MCIF's SQA, Setup 2 yields stronger performance on ASR and ST.
\section{Submitted Model}\label{sec:submission}

We select our model based on its MCIF performance. Overall, we were unable to produce a single run that achieves the best score across all tasks, as improvements in one task tend to degrade performance in others. Therefore, we prioritize SQA as our primary task and use it as the key criterion for model selection. Based on Table~\ref{tab:mcif_results}, we submit setup~1 as our main submission. We also resubmit BEST-IWSLT-IF as a contrastive, unconstrained system. Our competition results are discussed in Appendix Section~\ref{appendix:iwslt2026}.

\section{Conclusion}\label{sec:conclusion}

In this paper, we presented NLE’s submission to the instruction-following speech processing short track at IWSLT 2026 under the constrained setting. We developed multimodal models capable of jointly performing ASR, ST, and SQA from English speech into Chinese, German, and Italian. Building on our previous pipeline~\cite{lee-etal-2025-naver}, we replaced the transformer-based speech projector with an updated \speechmapper{} projector~\cite{speechmapper}. We also introduced a synthetic dataset of scientific talks, \textit{fakACL}, to mitigate domain mismatch between training and evaluation. Despite relying on a substantially smaller LLM backbone, our final system outperforms last year’s best submission in the short track.

{
    \small
    \bibliographystyle{ieeenat_fullname}
    \bibliography{custom}
}

\clearpage
\appendix
\clearpage
\section{Appendix}


\subsection{Qwen-based SQA Invalid Split}\label{sec:qweninvalid}

We aim to generate high-quality invalid questions for training our speech LLM. To ensure this quality, the generated questions must be unrelated to the speech content. Since LibriSQA is an audiobook-based dataset, we therefore focus on creating questions grounded in broad conversational topics.

We first prompted \qwen{} to produce a large set of conversation topics using: \begin{quote}
``\textit{List about 1000 conversation topics, without numbering or adding any comment of explanation.}''\end{quote} After post-processing the output, we retained 413 topics. Next, for each one, we prompted the LLM with: \begin{quote}``\textit{Generate 30 words related to \textbf{TOPIC}. Produce a comma-separated list of words, without any explanation.}''\end{quote} Then, using the resulting list, we finally prompted the model with: \begin{quote}``\textit{Ask 25 different questions about \textbf{WORD}. Do not repeat the question, include no comments, and output only the questions, one per line.}''\end{quote}

The output was automatically filtered to remove sentences that did not end with a question mark, resulting in 62,040 English questions. Some examples are shown below:
\begin{itemize}
    \item \textit{What role do animals play in traditional rites?}
    \item \textit{How does still air interact with thermal radiation?}
    \item \textit{Can a revoked loan be reapproved?}
    \item \textit{What is the role of the saw blade in determining cut quality?}
    \item \textit{How do social media platforms influence the perception of friendship and connection?}
    \item \textit{What is the significance of putrefaction in forensic science?}
\end{itemize}

\subsection{fakACL Creation}\label{sec:fakaclcreation}

Our goal in creating \textit{fakACL} is to mimic ACL60-60~\cite{salesky-etal-2023-evaluating} and MCIF. These datasets consist of oral presentations that typically begin with the authors introducing themselves and stating the title of the paper, followed by an overview of the main findings.

Our dataset creation process consists of three stages: (1) script creation, (2) segmentation and speech synthesis, and (3) QA generation. We now describe these stages.

\subsubsection{Script Creation}

The goal of this stage of the dataset creation process is to generate high-quality presentation scripts on NLP-related topics.
We started the process by first prompting \qwen{} to: \begin{quote}``\textit{List 60 sub-fields related to ACL/NLP conferences, one subfield par line.}''\end{quote} This resulted in a list of 56 ACL conference topics, with entries such as \textit{Healthcare NLP}, \textit{Legal document analysis} and \textit{Education}. We do not check for duplicated conference topics.

Next, we generated a collection of possible paper titles for each one of the topics. We prompted the LLM with: \begin{quote}``\textit{List 40 paper titles from ACL conferences related to: \textbf{SUB-FIELD}. Output only the paper titles, one per line. Nothing else.}''\end{quote} The result was a collection of 2,527 paper titles. Examples of generated paper titles were ``\textit{Phonological and Morphological Development in Multilingual Children}'' and ``\textit{Emotion-Driven Text-to-Speech with Multimodal Inputs}''.

Finally, we prompted the LLM with: \begin{quote}``\textit{Write a 12 sentences oral script for presenting the ACL paper entitled: \textbf{TITLE}.}''\end{quote} After manual processing, this resulted in 2,497 scripts. An example of a generated script is given at Table~\ref{tab:synthetic_script_example}.

\begin{table*}[t]
\centering
\small
\renewcommand{\arraystretch}{1.1}

\begin{tabularx}{\textwidth}{X}
\toprule
\textbf{Synthetic Script from \textit{fakACL}} \\
\midrule

Good morning, everyone. Today, I'm excited to present our paper titled
Cross-Modal Retrieval using Cross-Modal Semantic Matching and Contextual Awareness.
In traditional retrieval systems, matching content across different modalities
like text and images has been challenging due to inherent semantic gaps.
Our approach tackles this by introducing a novel framework that combines
cross-modal semantic matching with contextual awareness.

We leverage deep neural networks to learn rich, shared semantic representations
between modalities, ensuring that queries from one modality can accurately
retrieve relevant content from another. Crucially, we incorporate contextual
awareness to understand the environment and user intent such as time, location,
or user preferences before generating retrieval results.

This allows the system to go beyond literal matches and deliver more relevant,
human-like responses. We validate our method on benchmark datasets including
CORD-19 and MS-COCO, demonstrating significant improvements over baseline
models in precision and recall.

Our experiments show that contextual signals improve retrieval performance
by up to 18\% in real-world scenarios. Additionally, we conduct ablation
studies that confirm the importance of both semantic matching and contextual
components.

The framework is scalable and can be adapted to various applications from
medical image search to e-commerce product recommendations. We believe this
work bridges the gap between technical accuracy and real-world usability in
cross-modal systems.

Finally, we envision future work exploring dynamic context modeling and
multimodal feedback loops to further refine retrieval. Thank you for your
attention. I'd be happy to take any questions. 
\\\midrule
\textbf{Synthetic Questions and Answers from \textit{fakACL}} \\ \midrule
\textbf{Context from script:} ``Crucially, we incorporate contextual awareness to understand the environment and user intent such as time, location, or user preferences before generating retrieval results.''\\

\textbf{Question 1:} ``What does the system use to understand user intent?''

\textbf{Answer 1:} ``Contextual awareness including time, location, and user preferences.''\\

\textbf{Question 2:} ``How does the system enhance retrieval results?''

\textbf{Answer 2:} ``By aligning retrieval with time, location, and user preferences.''

\\\bottomrule
\end{tabularx}
\caption{Example of a synthetic presentation script and QA produced by \qwen{}.}
\label{tab:synthetic_script_example}
\end{table*}

\subsubsection{Segmentation and Speech Synthesis}

After the scripts are created, the next step consists of segmenting the content into blocks to be synthesized by \seamless{}. This stage includes preprocessing steps such as number and symbol normalization (e.g., ``30\%'' becomes ``thirty percent''), as well as text normalization.

We use the Python library \texttt{spaCy} to split the scripts into sentences, normalize the text, and discard sentences containing fewer than 10 characters or fewer than 6 words. This process results in 21,400 sentences. We then synthesize the sentences using \seamless{} TTS with random speakers. The average utterance length is 8.6 seconds.

\subsubsection{QA Generation}

Lastly, after obtaining the segmented and normalized scripts, our next goal was to generate valid question–answer pairs. For this stage, we discarded the first three and the last two sentences of each script, as these tended to be overly generic and usually corresponded to greetings or acknowledgments.

For each text segment, we prompted \qwen{} with: \begin{quote}``\textit{Generate 2 pairs of question and very short answer. Generate the question in one line, the answer in the next line, prefix the question by ``Q:'' and the answer by ``A:''. Generate strictly and solely based on the content: \textbf{CONTENT}.}''\end{quote}
We automatically filtered the outputs based on the question–answer prefixes and subsequently performed manual verification to remove clear hallucinations. This process yielded a final set of 38,968 questions.

\subsection{IWLST 2026 IF Short Results}\label{appendix:iwslt2026}

Table~\ref{tab:iwslt2026} presents the official results for our primary (Setup~1 from Table~\ref{tab:results1}) and contrastive (BEST-IWSLT25-IF) submissions. Overall, the results confirm that we were able to build a stronger system than last year's submission, despite relying on a smaller and weaker LLM backbone.

Our primary submission outperforms BEST-IWSLT25-IF across all languages for ST, while lagging by only 0.2\% WER on ASR. For QA, this year's model again improves performance across all languages except Italian, for which we found the backbone to be particularly weak. 

Finally, the inclusion of MCQ training allowed our model to perform the surprise task of Quality Estimation~(QE-accuracy) while respecting the proposed zero-shot prompt~(QE-format-accuracy). In contrast, last year's solution struggled to follow the required format and often generated answers instead of selecting options, making it incapable of correctly answering cases where the correct option did not correspond to the language of the prompt.

\begin{table*}[t]
\centering
\small
\setlength{\tabcolsep}{6pt}

\resizebox{\linewidth}{!}{%
\begin{tabular}{llccccc}
\toprule
\textbf{Model} & \textbf{Lang} & \textbf{TRANS-COMET} & \textbf{QA-BERTScore} & \textbf{QE-accuracy} & \textbf{QE-format-accuracy} & \textbf{ASR-WER} \\
\midrule
NLE\_IWSLT26\_IF\_SHORT\_CONSTRAINED\_PRIMARY
& en & -- & \textbf{0.531} & -- & -- & 0.136 \\
NLE\_IWSLT26\_IF\_SHORT\_CONSTRAINED\_PRIMARY
& it & \textbf{0.763} & 0.456 & -- & -- & -- \\
NLE\_IWSLT26\_IF\_SHORT\_CONSTRAINED\_PRIMARY
& de & \textbf{0.765} & \textbf{0.470} & \textbf{0.786} & \textbf{0.997} & -- \\
NLE\_IWSLT26\_IF\_SHORT\_CONSTRAINED\_PRIMARY
& zh & \textbf{0.794} & \textbf{0.487} & \textbf{0.894} & \textbf{1.000} & -- \\

\midrule
NLE\_IWSLT26\_IF\_SHORT\_UNCONSTRAINED\_CONSTRASTIVE
& en & -- & 0.501 & -- & -- & \textbf{0.134} \\
NLE\_IWSLT26\_IF\_SHORT\_UNCONSTRAINED\_CONSTRASTIVE
& it & 0.733 & \textbf{0.514} & -- & -- & -- \\
NLE\_IWSLT26\_IF\_SHORT\_UNCONSTRAINED\_CONSTRASTIVE
& de & 0.749 & 0.462 & 0.333 & 0.005 & -- \\
NLE\_IWSLT26\_IF\_SHORT\_UNCONSTRAINED\_CONSTRASTIVE
& zh & 0.755 & 0.466 & 0.500 & 0.014 & -- \\
\bottomrule
\end{tabular}%
}
\caption{Official results for our systems submitted to IWSLT 2026 IF short track.}
\label{tab:iwslt2026}
\end{table*}

\begin{table}
\centering
\resizebox{\columnwidth}{!}{
\begin{tabular}{c|ccc}\toprule
\multicolumn{1}{c}{\textbf{Dataset}}  & \textbf{Task}                    & \textbf{Language} & \multicolumn{1}{l}{\textbf{\# Samples}} \\\midrule
\multirow{3}{*}{\textbf{CoVoST2}}     
& \textbf{ASR}                     
& en                
& 289,413\\ 
\cline{2-4} 
& \multirow{2}{*}{\textbf{ST/MT}}
& en-de
& 289,413\\
&
& en-zh
& 289,413\\\midrule
\multirow{3}{*}{\textbf{EuroParlST}}
& \textbf{ASR}
& en
& 35,372
\\\cline{2-4} 
& \multirow{2}{*}{\textbf{ST/MT}}
& en-de
& 32,628\\
&
& en-it
& 29,552
\\\midrule
\multirow{4}{*}{\textbf{GigaST}}
& \textbf{ASR}
& en
& 7,645,641
\\\cline{2-4} 
& \multirow{3}{*}{\textbf{ST/MT}}
& en-de
& 7,645,641\\
&
& en-it*
& 3,004,804\\
&
& en-zh
& 7,645,641\\\midrule
\multirow{11}{*}{\textbf{LibriSQA}}
& \textbf{ASR}                     
& en                
& 104,014\\\cline{2-4} 
& \multirow{3}{*}{\textbf{MT}}
& en-de*
& 36,707\\
&
& en-it*
& 75,378\\
&
& en-zh*
& 22,211\\\cline{2-4} 
& \multirow{4}{*}{\textbf{SQA/QA (valid)}}
& en
& 208,038\\
&
& en-de*
& 59,274\\
&
& en-it*
& 127,295\\
&
& en-zh*
& 29,385\\\cline{2-4} 
& \multirow{4}{*}{\textbf{SQA/QA (invalid)}}
& en
& 169,068\\
&
& en-de*
& 79,578\\
&
& en-it*
& 123,632\\
&
& en-zh*
& 58,092\\\midrule
\multirow{11}{*}{\textbf{fakACL}}
& \textbf{ASR}                     
& en$\dag$                
& 21,400\\\cline{2-4} 
& \multirow{3}{*}{\textbf{ST/MT}}
& en$\dag$-de*
& 9,722\\
&
& en$\dag$-it*
& 14,999\\
&
& en$\dag$-zh*
& 4,659\\\cline{2-4} 
& \multirow{4}{*}{\textbf{SQA/QA (valid)}}
& en$\dag$
& 38,968\\
&
& en$\dag$-de*
& 8,763\\
&
& en$\dag$-it*
& 22,290\\
&
& en$\dag$-zh*
& 5,341\\\cline{2-4} 
& \multirow{4}{*}{\textbf{SQA/QA (invalid)}}
& en$\dag$
& 38,968\\
&
& en$\dag$-de*
& 21,192\\
&
& en$\dag$-it*
& 31,333\\
&
& en$\dag$-zh*
& 19,549\\
\bottomrule                            
\end{tabular}}
\caption{Training sets statistics by task. For ST/MT sets, target side is duplicated. * denotes synthetic text obtained via \seamless{} MT; $\dag$ indicates splits generated with \seamless{} TTS. For fakACL, all English textual content is produced via \qwen prompting.}
\label{tab:trainingdata}
\end{table}

\begin{table*}
\centering
\resizebox{\textwidth}{!}{
\begin{tabular}{lccc|ccc|ccc|ccc|cccc}\toprule
& \multicolumn{3}{c}{\textbf{CoVoST}}      & \multicolumn{3}{c}{\textbf{EuroParlST}} & \multicolumn{3}{c}{\textbf{GigaST}}   & \multicolumn{3}{c}{\textbf{LibriSQA}} & \multicolumn{4}{c}{\textbf{fakeACL}}\\
& \textbf{ASR} & \textbf{ST} & \textbf{MT} & \textbf{ASR} & \textbf{ST} & \textbf{MT} & \textbf{ASR} & \textbf{ST} & \textbf{MT} & \textbf{ASR} & \textbf{MT} & \textbf{SQA/QA} & \textbf{ASR} & \textbf{ST} & \textbf{MT} & \textbf{SQA/QA} \\\midrule
\textbf{A} \speechmapper{} (ASR)   
& $$ \ding{51} $$  & $$ \ding{55} $$  & $$ \ding{55} $$   
& $$ \ding{51} $$  & $$ \ding{55} $$  & $$ \ding{55} $$ 
& $$ \ding{51} $$  & $$ \ding{55} $$  & $$ \ding{55} $$ 
& $$ \ding{51} $$  &  $$ \ding{55} $$ &  $$ \ding{55} $$  
&  $$ \ding{55} $$ &  $$ \ding{55} $$ &  $$ \ding{55} $$ & $$ \ding{55} $$ \\
\textbf{B} Text-only LoRA (MT/QA)
& $$ \ding{55} $$  & $$ \ding{55} $$  & $$ \ding{51} $$   
& $$ \ding{55} $$  & $$ \ding{55} $$  & $$ \ding{51} $$ 
& $$ \ding{55} $$  & $$ \ding{55} $$  & $$ \ding{51} $$ 
& $$ \ding{55} $$  &  $$ \ding{51} $$ &  $$ \ding{51} $$  
&  $$ \ding{55} $$ &  $$ \ding{55} $$ &  $$ \ding{51} $$ & $$ \ding{51} $$\\
\textbf{C} Multimodal model (ASR/ST/MT/SQA/QA)
& $$ \ding{51} $$  & $$ \ding{51} $$  & $$ \ding{51} $$
& $$ \ding{51} $$  & $$ \ding{51} $$  & $$ \ding{51} $$
& $$ \ding{55} $$  & $$ \ding{55} $$  & $$ \ding{51} $$
& $$ \ding{55} $$  &  $$ \ding{51} $$ &  $$ \ding{51} $$  
& $$ \ding{55} $$ &  $$ \ding{55} $$ &  $$ \ding{51} $$ & $$ \ding{51} $$\\
\bottomrule                  
\end{tabular}}
\caption{List of datasets and splits used for training our presented models. Statistics are presented in Table~\ref{tab:trainingdata}.}
\label{tab:listdatasets}
\end{table*}

\begin{table*}[]
\centering
\begin{tabular}{p{0.25\textwidth} p{0.7\textwidth}}\toprule
\textbf{}        & \multicolumn{1}{c}{\textbf{User Prompt}}\\\midrule
\textbf{Speech Prefix} & Content: \textless{}speech\textgreater{}{[}\textbf{SPEECH EMBEDDINGS}{]}\textless{}/speech\textgreater{}\textbackslash{}n\\\midrule
\textbf{Text Prefix} & Content: \textless{}text\textgreater{}{[}\textbf{SPEECH TRANSCRIPTION}{]}\textless{}/text\textgreater{}\textbackslash{}n\\\midrule
\textbf{ASR}& \begin{tabular}[c]{@{}l@{}}Question: Can you transcribe the Speech content into English text?\textbackslash{}n\end{tabular}\\\midrule
\textbf{ST/MT (de)}& Question: Können Sie den Inhalt der Rede in den deutschen Text übersetzen?\textbackslash{}n\\
\textbf{ST/MT (it)}& Question: Puoi tradurre il contenuto del discorso in testo italiano?\textbackslash{}n\\
\textbf{ST/MT (zh)}& Question: \begin{CJK*}{UTF8}{gbsn}你能把演讲内容翻译成中文吗?\textbackslash{}n\end{CJK*}\\
\midrule
\textbf{SQA/QA \small{(Free format)}}& \begin{tabular}[c]{@{}l@{}}Question: {[}\textbf{QUESTION}{]}\textbackslash{}n\end{tabular}\\
\textbf{SQA/QA \small{(MCQ)}}& \begin{tabular}[c]{@{}l@{}}Question: {[}\textbf{QUESTION+OPTIONS}{]}\textbackslash{}n\end{tabular}\\\midrule
\textbf{Surprise Task}& \begin{tabular}[c]{@{}l@{}}Question: {[}\textbf{SURPRISE INSTRUCTION}{]}\textbackslash{}n\end{tabular}\\\midrule
\textbf{Suffix} & \textbackslash{}nYour answer: \\
\textbf{Suffix (zero-shot ASR)} & Do not add anything else to your answer. \textbackslash{}nYour answer: \\
\textbf{Suffix (zero-shot ST de)} & Antworten Sie nur mit der Übersetzung. Fügen Sie Ihrer Antwort nichts weiter hinzu.\textbackslash{}nYour answer: \\
\textbf{Suffix (zero-shot ST it)} & Rispondi solo con la traduzione. Non aggiungere altro alla tua risposta.\textbackslash{}nYour answer:\\
\textbf{Suffix (zero-shot ST zh)} & \begin{CJK*}{UTF8}{gbsn}仅回复翻译。请勿在答案中添加任何其他内容。\end{CJK*}\textbackslash{}nYour answer:\\
\bottomrule
\end{tabular}
\caption{The user turn prompt template used for training and/or evaluating models. For speech tasks, the user prompt is given by \textbf{Speech Prefix+Task+Suffix}, for textual tasks, the user prompt is given by \textbf{Text Prefix+Task+Suffix}. \textbf{ST/MT instructions} were obtained by translating the instruction ``\textit{Can you translate the Speech content into \textbf{{[}German/Italian/Chinese{]}} text?}'' and ``\textit{Answer only with the translation. Do not add anything else to your answer.}'' to corresponding target languages using \seamless{}. \textbf{SQA/QA instructions} are either \textit{free format}~(direct answer) or \textit{multiple-choice questions} (MCQ). In the case of MCQ, options are included in the same line of the question, and both are translated to the target language in case of multilingual SQA. \textbf{Surprise task instructions} are included after replacing the ``\textbackslash{}n'' by spaces.}
\label{tab:prompts}
\end{table*}
\begin{table*}
\resizebox{\textwidth}{!}{
\begin{tabular}{p{0.48\textwidth} p{0.48\textwidth}}\toprule
\textbf{Reference} & \textbf{Generation} \\\midrule
\fix{This} is a joint work with John \fix{Gauthier}, Aaron Mueller, Kanishka Misra, Karen \fix{Fences}, Roger \fix{Levy}, and \fix{Adina Williams.}
&
\fix{It'}s a joint work with John \fix{Gautier}, Aaron Mueller, Kanishka Misra, Karen \fix{Fentos}, Roger \fix{Le}, and \fix{Andina} \fix{Will. Will. Will.}
\\\midrule
Mr President, what worries me most is the suggestion that the European Investment Bank has a significant role to play in solving the financial crisis – \fix{that it can} somehow act as the Europe-wide body to provide stimulus where Member States have singularly failed to coordinate their own stimuli. & Mr. President, what worries me most is the suggestion that the European Investment Bank has a significant role to play in solving the financial crisis. \fix{It is claimed that the bank could} somehow act as the Europe-wide body to provide stimulus where member states have singularly failed to coordinate their own stimuli. \\\midrule
From the experience I have as a member of the national parliament and \fix{talking} with many people \fix{for} a long time, I would really be in \fix{favour of} smart sanctions – \fix{targeted}, for example, \fix{at} specific members of the Revolutionary Guard, \fix{putting them on a veto list for visits, or at other specific persons.}
&
From the experience I have as a member of the national parliament and \fix{having discussed this} with many people \fix{over} a long time, I would really \fix{support} smart sanctions \fix{targeting}, for example, specific members of the Revolutionary Guard, \fix{in order to target or punish specific individuals.}
\\\midrule
C. Valor and fortune
&
The speech states: *``each party hurrying to a battle where value and and fortune chiefly the the success.''*  \textbackslash{}n\textbackslash{}nThis indicates that the primary factors determining battle success were **valor** (bravery) and **fortune** (luck).\textbackslash{}n\textbackslash{}nLooking at the options:\textbackslash{}n\textbackslash{}n- A. Experience and knowledge → not mentioned  \textbackslash{}n- B. Strategic planning → not mentioned  \textbackslash{}n- C. Valor and fortune → directly supported by the speech  \textbackslash{}n- D. Ignorance and impetuous conduct → mentioned as something historians may have overlooked, not as a factor of success  \textbackslash{}n\textbackslash{}n**Correct answer: C. Valor and fortune** 
\\
\bottomrule
\end{tabular}}
\caption{Some examples of generations for \qwen{} using \speechmapper{}~(A) embeddings. In the top, an example of hallucination related to unknown named entities at MCIF; in the middle two examples of ASR rephrasing with EuroParlST; and in the bottom an example of extremely verbose SQA output for LibriSQA.}
\label{tab:generationsm}
\end{table*}

\end{document}